# Deep Learning Segmentation of Diffusion-Weighted MRI Acute Ischaemic Stroke: A Pragmatic Evaluation Across Three Datasets

Atle Bjørnerud[1,8], Till Schellhorn[1,2], Thor H. Skattør[2,7], Terje Nome[2], Jon André Ottesen[1], Anne Hege Aamodt[5,6], Bradley J MacIntosh[1,3,4]

1) Computational Radiology & Artificial Intelligence Unit, Department of Physics and Computational Radiology, Oslo University Hospital, Oslo, Norway

2) Division of Radiology and Nuclear Medicine, Oslo University Hospital, Oslo, Norway

3) Department of Medical Biophysics, Faculty of Medicine, University of Toronto, Canada
4) Hurvitz Brain Sciences, Sandra Black Centre for Brain Resilience & Recovery, Physical Sciences Platform, Sunnybrook Research Institute, Canada

5) Department of Neurology, Oslo University Hospital, Oslo, Norway

6) Department of Neuromedicine and Movement Science, The Norwegian University of Science and Technology, Trondheim, Norway

7) Institute of Clinical Medicine, Faculty of Medicine, University of Oslo, Oslo, Norway

8) Center for Lifespan Changes in Brain and Cognition, Department of Psychology, University of Oslo, Oslo, Norway

**Corresponding author:**

Atle Bjørnerud, PhD. Computational Radiology & Artificial Intelligence Unit, Oslo University Hospital, Oslo, Norway. Email: atlebjo@uio.no

## Abstract

**Objective:** Diffusion-weighted imaging (DWI) Magnetic Resonance Imaging (MRI) is the gold standard to visualize and quantify acute ischaemic stroke (AIS). Although numerous studies demonstrate the capability of deep learning to segment the AIS lesions, further research is warranted to identify optimal image inputs and the choice of the deep learning model. The aim of this study was to evaluate whether accurate acute stroke lesion segmentation can be achieved using a pragmatic deep learning approach requiring minimal preprocessing and clinically feasible inference time.

**Materials & Methods:** We trained self-configured nnU-Net models using n=1744 DWI cases from local, national, and open-access data sources. Models were tested on n=436 test sample. The models were evaluated with five-fold cross validation for four experimental conditions: brain extracted (no bet versus +bet), and use of DWI as a single input or DWI and the apparent diffusion coefficient image as two channel image inputs. We considered two architectures (i.e. baseline nnU-Net [base] or a larger residual encoder nnU-Net [ResEnc]). Segmentation results were compared against the DeepISLES ensemble model from the 2022 ISLES challenge.

**Results:** In the test set (n=436), median (interquartile range) Dice similarity coefficient (DSC) for the base model was 0.84 (0.19). For the base model, we observed significant for two out of six pairwise comparisons for the choice of model inputs. The ResEnc model training led to small but appreciable segmentation differences compared to the base model. Namely, ResEnc produced a significantly higher Dice compared to the base model when considering DWI, DWI+bet, and DWI+ADC configurations (all $p < 0.02$), but not for DWI+ADC+bet ($p > 0.50$). The base model performed significantly better than the DeepISLES model, notably in the stroke cohort characterized by smaller stroke volumes (Related-Samples Signed Rank test, $p < 0.01$).

**Conclusions:** The base nnU-Net trained with only DWI as the input and no preprocessing enabled fast and accurate AIS lesion segmentation. The methods development may facilitate clinical research in acute stroke imaging workflows.

## Introduction

Stroke remains a major burden of disease, with an estimated 12 million stroke events occurring annually (1) . While computed tomography (CT) remains the dominant modality for initial acute stroke triage due to its widespread availability, magnetic resonance imaging (MRI) is considered the gold-standard to visualize acute ischaemic stroke (AIS) using a diffusion-weighted imaging (DWI) acquisition. The decision to perform MRI for the acute stroke imaging depends on several factors that include local resources and clinical presentations, e.g. wake-up stroke, small vessel disease, or recurrent stroke / transient ischaemic attack (2,3). Acute stroke MRI offers inherent advantages due to the superior tissue contrast compared to CT, but it is not without its challenges (4,5).

Deep learning has emerged as a powerful tool in radiology for image analysis and decision support. For instance, segmenting the AIS lesion on DWI is a topic of intense research interest, with the goal of robust and accurate automated lesion delineation (6). Numerous studies and international challenges demonstrate strong segmentation performance across large and heterogeneous datasets (7), underlining the potential of these methods to support quantitative imaging, prognostication, and evaluation of treatment effects in acute stroke. Despite these advances, translation of DWI segmentation models into routine clinical radiology remains limited, partly due to methodological complexity and workflow-related challenges (8).

Echo planar imaging (EPI) is one of the most widely used methods for rapid imaging, known for the efficient rastering through MRI k-space, and the acquisition approach used in DWI (9). Despite advanced imaging approaches EPI fundamentally suffers from spatial and intensity distortions relative to structural MRI pulse sequences.

An ongoing image analysis challenge is the need for high-quality AIS segmentation that rely on as few DWI image inputs and preprocessing steps as possible. Most approaches for AIS lesion segmentation rely on a skull-stripping step for brain extraction and/or reliance on additional input images like the apparent diffusion coefficient (ADC) maps (7,10,11). For research purposes, these requirements are reasonable but add to the complexity of the imaging data pipeline. For instance, the brain extraction step can easily be evaluated with visual inspection, but this is impractical in a resource limited clinical setting. Although deep learning-based brain extraction tools such as HD-BET perform well on high-resolution structural MRI (12), their performance is expected to reduce or exhibit more variability when considering DWI. Degraded segmentation performance is attributed to poorer image quality, geometric distortions, susceptibility artefacts, and a lack of representative training data. As a result, structural MRI sequences are frequently incorporated to improve brain extraction, necessitating additional co-registration steps that increase computational complexity and processing time (7).

The requirement for both DWI and ADC images introduces further challenges in clinical practice. Depending on the MRI vendor settings and hospital PACS configurations, ADC maps may be included within the same dataset or provided as a separate series. Automated identification and handling of the correct DWI and ADC inputs as part of a radiology-integrated pipeline is non-trivial due to limited standardization across systems. These variations add complexity to deployment and introduce multiple potential sources of error, reducing robustness in real-world clinical environments.

The current study is motivated by the need to evaluate the AIS segmentation performance across locally/regionally collected stroke imaging datasets, aiming for a pragmatic solution with minimal preprocessing, one that avoids image alignment, and conducive to rapid inference directly from routinely acquired DWI. Such an approach could lower barriers to clinical implementation and improve integration into existing radiological workflows. The self-configuring nnU-Net framework has demonstrated state-of-the-art performance across a wide range of biomedical image segmentation tasks, including stroke lesion segmentation, as reflected by top-ranking models in the ISLES'22 ischaemic stroke segmentation challenge (7). The aim of this study was to evaluate whether accurate acute stroke lesion segmentation can be achieved using a simplified, clinically feasible deep learning approach with minimal preprocessing requirements and acceptable inference times for routine radiological use.

## Material and Methods

### Datasets

In this retrospective study, MRI was obtained from three separate AIS studies; Nor-COAST, OSCAR, and the Stroke Outcome Optimization Project (SOOP). The former two are national study cohorts that received approval from respective regional ethics committees (REK 2015/1844 for OSCAR and 2015/171 for Nor-COAST). Collection and distribution of the open access data described in the SOOP repository was approved under protocol Pro00078716 of the Prisma Health Committee (13). MRI data consisted of DWI, derived ADC maps, and fluid attenuated inversion recovery (FLAIR) images. Data were provided and processed in Neuroimaging Informatics Technology Initiative (NIfTI) format. For all three datasets, AIS lesions were annotated on the high b-value DWI in the imaging series per patient. Twenty percent of the sample was held out as the test set and the remaining data were used in an 80/20 split for training and validation samples.

*The Norwegian Cognitive Impairment after Stroke Study* (Nor-COAST, NCT02650531) is a multicenter (five participating hospitals), prospective observational cohort study with consecutive inclusion during the acute phase and with follow-up at three and 18 months, and at three years (14). Only AIS cases from the acute phase were included in the current study.

*The Oslo Acute Reperfusion Stroke Study* (OSCAR, NCT06220981) is a prospective observational cohort study on acute ischaemic stroke patients who received endovascular thrombectomy (EVT) at Oslo University Hospital between 2017 and 2022. MRI was acquired before EVT and repeated 12-36 h after EVT. The OSCAR inclusion criteria and protocol are described in detail elsewhere (15).

*The Stroke Outcome Optimization Project* (SOOP) is a large public dataset of annotated clinical MRIs and metadata of patients with acute stroke (13), available through the OpenNeuro portal (https://openneuro.org).

**Data exclusion criteria**

Patients were excluded based on intracerebral haemorrhage, excessive head motion in DWI, missing ground truth masks, or corrupted data. Intracerebral hemorrhage cases were excluded through chart review and facilitating by taking the average DWI intensity of the GT annotated lesion and comparing against the normal appearing brain parenchyma.

**Ground truth annotations**

Ground truth (GT) DWI stroke lesion segmentations were performed by experienced neuroradiologists in all three datasets as follows:

*Nor-COAST*: AIS lesions were labelled from the DWI scans by a senior neuroradiologist (T.S.) using the snake tool in ITK-SNAP for semi-automatic annotations (16).

*OSCAR:* Independent segmentation of hyper-intense DWI lesions was performed on a subset of 70 consecutive cases by two experienced neuroradiologists (T.H.S., T.N.). Inter-rater reliability, measured by the Dice coefficient, was high with a mean value of 0.86. Discrepancies were resolved by consensus when the Dice coefficient was below 0.70. Following consensus, one rater (T.H.S) segmented the remaining cases and these segmentation masks were used as ground truths (15). Segmentation was performed in the nordicICE software package (https://crai.no/product/nordicice).

*SOOP:* Three trained neuroscientists used the MRIcroGL software (17) to manually inspect and trace hyperintense DWI lesions. A similar process was used to identify participants that additionally showed evidence of chronic stroke lesions, depicted as hypointense on the same ADC images (13). Acute and chronic stroke lesion files are stored separately in the SOOP OpenNeuro database, and only acute lesions were included in the current study.

**Model architecture**

Using the self-configuring nnU-Net framework (18), we conducted a series of experiments based on the following configurations: (1) only DWI (DWI), (2) DWI with brain extraction (DWI+bet), (3) including the ADC images (DWI+ADC), and (4) DWI and ADC with brain extraction (DWI+ADC+bet). The resulting four configurations were trained using two model configurations: a standard 3D nnU-Net full resolution architecture with the default plan obtained from the auto-configuration stage (base model), an nnU-Net architecture with residual connections in the encoder (ResEnc). Specifically, the ResEnc large configuration referred was the standard option in work reported by others (19). It required approximately 3 times more GPU memory than the base model, and approximately 4 times longer runtime as stated in a recent comparison (20). Both models were run with the default auto-configurations, except for number of epochs which was changed from the default value of 1000 to 500.
Model configurations are detailed in Table 1. Compute times for training and predictions were recorded and compared between models.

**Table 1.** Main parameters for the two nnU-Net configurations that were trained using local and open-source data sources.

| Model parameter | Base model: nnU-Net | nnU-Net ResEnc(L) |
|---|---|---|
| Number of down-sampling stages | 6 | 6 |
| Number of features / stage | [32,64,128,256,320,320] | [32,64,128,256,320,320] |
| Number of convolution blocks /stage (encoder) | 2 | NA |
| Number of transposed convolution blocks/stage (decoder) | 2 | 1 |
| Number of residual blocks/stage (encoder) | NA | [1,3,4,6,6,6] |
| Patch size | 32x256x192 | 32x256x192 |
| Kernel size | [1,3,3],[1,3,3],[3,3,3], [3,3,3],[3,3,3],[3,3,3] | [1,3,3],[1,3,3],[3,3,3], [3,3,3],[3,3,3],[3,3,3] |
| Strides /stage | [1,1,1], [1,2,2], [1,2,2], [2,2,2], [2,2,2], [2,2,2] | [1,1,1], [1,2,2], [1,2,2], [2,2,2], [2,2,2], [2,2,2] |
| Number of trainable parameters (x $10^6$) | 87.5 | 381.3 |

See text for additional model descriptions.

The two in-house trained nnU-Net models were compared against an external model called DeepISLES, which was clinically validated for AIS segmentation (7). DeepISLES is an ensemble of the three best performing models from the Ischemic Stroke Lesion Segmentation Challenge (https://www.isles-challenge.org/). This ensemble consisted of a majority voting strategy such that the binary output mask (lesion: yes/no) was determined by a positive prediction from at least two of the three models. DeepISLES incorporated different model architectures: SEALS, which is based on the nnU-Net, NVAUTO based on SegResNet (21), and SWAN using the Factorizer algorithm (22). DeepISLES required a brain extracted DWI and the use of ADC and FLAIR images as three channel inputs. The brain extraction preprocessing relied on HD-BET that was applied to the FLAIR, followed by co-registration of the FLAIR to DWI/ADC to apply the brain mask to the diffusion series. The same method was applied for the base and ResEnc configurations that relied on brain extracted DWI / ADC series. The DeepISLES model was accessed from Github and run via python script after local installation (https://github.com/ezequieldlrosa/DeepIsles; accessed 2026 March 20).

All nnU-Net configurations were trained with 5-fold cross-validation. In inference, the five models from cross-validation were used as an ensemble by averaging the softmax predictions. Model training and testing were performed on a single NVIDIA A100-PCIE-40GB GPU. The overall analysis workflow is summarized in Figure 1.

**Model performance evaluation**

In compliance with the ISLES'22 evaluation metrics, segmentation performance was evaluated in terms of Dice similarity coefficient (DSC), lesion-wise F1-score, absolute lesion volume difference (AVD) (in mL), and absolute lesion count difference (ALD) between ground truth and predicted lesions (7). The python script provided for the ISLES'22 challenge was used for all metrics evaluations.

**Statistical analysis**

Based on the Kolmogorov-Smirnov test, all performance metrics were found not to be normally distributed, hence non-parametric tests are reported throughout. Performance scores are reported as median +/- interquartile range (IQR). Friedman Two-Way Analysis of Variance by Ranks was used for estimating overall performance differences between the eight configurations of the in-house trained nnU-Net models and Wilcoxon signed rank test was used post-hoc for case-wise comparisons. Comparison between nnU-Net and DeepISLES was also tested using case-wise related-samples Wilcoxon signed rank test. The performance comparison was made separately for the three test sets (i.e. SOOP, OSCAR, Nor-COAST) to identify dataset specific differences in performance. In addition, differences in segmentation performance between nnU-Net and DeepISLES were compared according to GT lesion volume ranges: < 5 mL, 5-20 mL and > 20 mL, in accordance with the analysis described previously by de la Rosa et al (7). Independent-samples Kruskal-Wallis test was used to assess differences in overall GT-lesion volumes and number of isolated lesions between the three datasets included in the test set. Correlation between GT- and estimated lesion volumes were evaluated from linear regression and Pearson correlation analysis. Bland-Altman plots were generated to identify any bias in stroke volume estimates for the different models versus GT, A two-sided significance threshold of $p = 0.05$ was used for all tests after Bonferroni correction for multiple comparisons where appropriate.

Statistical analysis was performed in IBM SPSS statistics (version 31).

## Results

### Data Characteristics

The three AIS datasets consisted of a total of 2309 MRI cases, where 129 cases were excluded according to the exclusion criteria, leaving 2180 AIS cases for model training/validation (n=1744) and testing (n=436). For the comparison to DeepISLES, 12 cases from OSCAR test set were excluded due to missing FLAIR data. Data inclusion and analysis pipelines are summarized in Figure 1.

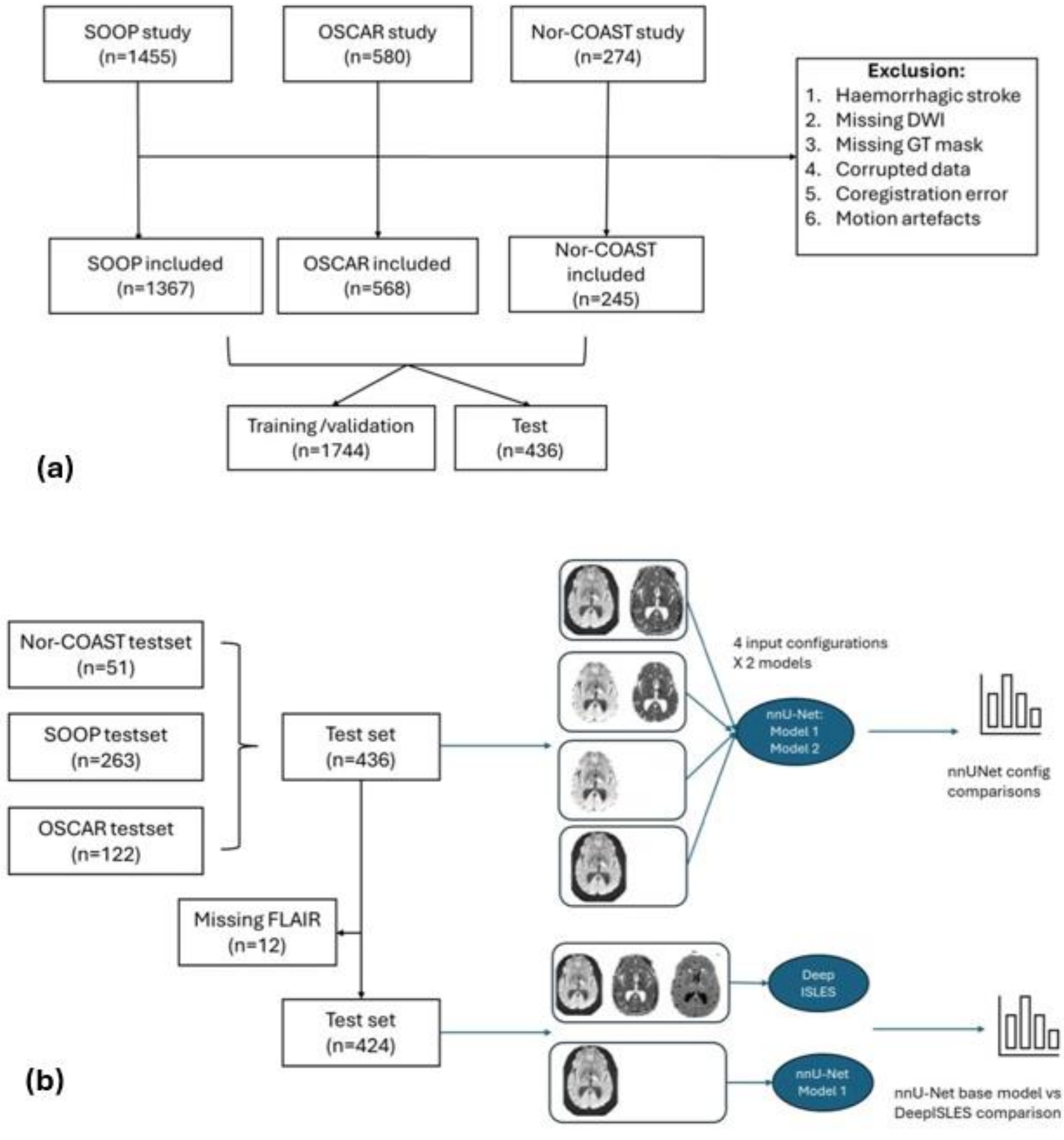

**Figure 1.** Summary of datasets and data exclusion (a) and analysis pipeline on the test sample (b). For the in-house nnU-Net implementations, four different input configurations were trained using two different nnU-Net architectures. The resulting eight configurations are then compared on the test set. In addition, the base model (DWI only) prediction is compared to the DeepISLES ensemble model which required ADC, DWI and FLAIR inputs. SOOP = Stroke Outcome Optimization Project. OSCAR= Oslo Acute Reperfusion Stroke Study and Nor-COAST = Norwegian Cognitive Impairment after Stroke Study.

**DWI lesion volumes**

The median (IQR) lesion volume (mL) from the GT segmentations were 9.4 (34.8), 17.2 (40.6) and 5.1 (13.3) mL, respectively for the SOOP, OSCAR and Nor-COAST datasets. The OSCAR dataset had significantly larger lesion volumes relative to SOOP and Nor-COAST (both $p = 0.003$). The corresponding number of isolated lesions were 2 (5), 10 (13.5) and 1 (2), respectively. The difference in number of isolated lesions between all three datasets was significant ($p < 0.001$).

**Comparison of nnU-Net configurations**

Table 2 summarizes the performance metrices across all eight of the nnU-Net model training experiments. Overall, all models and configurations had similar performance metrics. The base model with DWI only and no brain extraction achieved a median (IQR) DSC of 0.84 (0.19), lesion-wise F1-score of 0.77 (0.43), AVD of 1.37 (3.96) mL, and ALD of 1 (3). Despite similar median DSC scores, the differences between models and configurations were statistically significant (Friedman's two-way ANOVA by ranks, $p<0.001$). Specifically, ResEnc had higher

**Table 2.** Comparison of performance metrics (median, IQR) for the four input configurations and two nnU-Net models.

| | Model 1: nnU-Net base | | | | Model 2: ResEnc(L) nnU-Net | | | |
|---|---|---|---|---|---|---|---|---|
| Model input | DSC ↑ | F1 ↑ | AVD ↓ | ALD↓ | DSC ↑ | F1 ↑ | AVD ↓ | ALD↓ |
| DWI + ADC | 0.84 (0.19) | 0.76 (0.43) | 1.36 (3.90) | 1.00 (3.00) | 0.85 (0.18) | 0.80 (0.40) | 1.33 (4.092 | 1.00 (3.00) |
| DWI | 0.84 (0.19) | 0.80 (0.43) | 1.34 (4.02) | 1.00 (3.00) | 0.85 (0.20) | 0.80 (0.38) | 1.28 (3.61) | 1.00 (3.00) |
| DWI+ADC+bet | 0.85 (0.19) | 0.80 (0.41) | 1.34 (3.68) | 1.00 (3.00) | 0.85 (0.18) | 0.80 (0.38) | 1.35 (3.36) | 1.00 (3.00) |
| DWI+bet | 0.84 (0.19) | 0.80 0.40) | 1.44 (4.11) | 1.00 (3.00) | 0.85 (0.19) | 0.80 (0.38) | 1.41 (3.60) | 1.00 (3.00) |

DWI+ADC= both DWI and ADC images used as model input, DWI=only DWI image used. ResEnc(L) = nnU-Net with residual encoder. bet = brain extraction applied to input images. DSC= Dice Similarity Coefficient, F1= Lesion-wise F1-score, AVD =absolute lesion volume difference (mL) and ALD = absolute lesion count difference. ↑↓ = higher/lower is better.

case-wise DSC scores compared to the base model for a subset of input configurations. There was no difference in DSC between models with only DWI versus models with DWI+ADC as inputs. For ResEnc with DWI+ADC as two channel inputs, there was a DSC trend in favour of applying brain extraction. There was no significant difference in any of the other metrics across all nnU-Net model and input configuration combinations. Test statistics for the different model configurations in terms of differences in DSC are summarized in Table 3.

Model training time (one-fold) was 13 hours for the base model and 55 hours for ResEnc, with corresponding inference times of ~ 5.0 s and ~ 11.0 s per case.

**Table 3.** Summary of the pairwise comparison for DSC scores.

| Model configuration | Comparison | Std. test statistics | p-value |
|---|---|---|---|
| nnU-Net base model | DWI vs DWI-bet | -0.84 | 1.0 |
| nnU-Net base model | DWI vs DWI+ADC | 0.78 | 1.0 |
| nnU-Net base model | DWI vs DWI+ADC+bet | -2.83 | **0.028** |
| nnU-Net base model | DWI+ADC vs DWI-bet | -0.77 | 1.0 |
| nnU-Net base model | DWI+ADC vs DWI+ADC+bet | -2.75 | 0.35 |
| nnU-Net base model | DWI-bet vs DWI-ADC+bet | -1.95 | 0.28 |
| Base vs ResEnc | DWI | -5.17 | **<0.001** |
| Base vs ResEnc | DWI+bet | -3.39 | **0.020** |
| Base vs ResEnc | DWI+ADC | -4.45 | **<0.001** |
| Base vs ResEnc | DWI+ADC-bet | -2.37 | 0.50 |

The top rows (white shading) show the comparisons between the image input options from within the same model. The bottom half of rows (grey shading) show the comparisons between the base and ResEnc models. ResEnc refers to the model with residual connections in the encoder. Std test statistics = case-wise related-samples Wilcoxon signed rank test: negative value indicates better DSC results for brain extracted input or model with residual encoder. p-value = Bonferroni-corrected significance value. There was no significant difference between any of the other performance metrics: Lesion-wise F1-score, Absolute lesion volume difference and Absolute lesion count difference (data not shown).

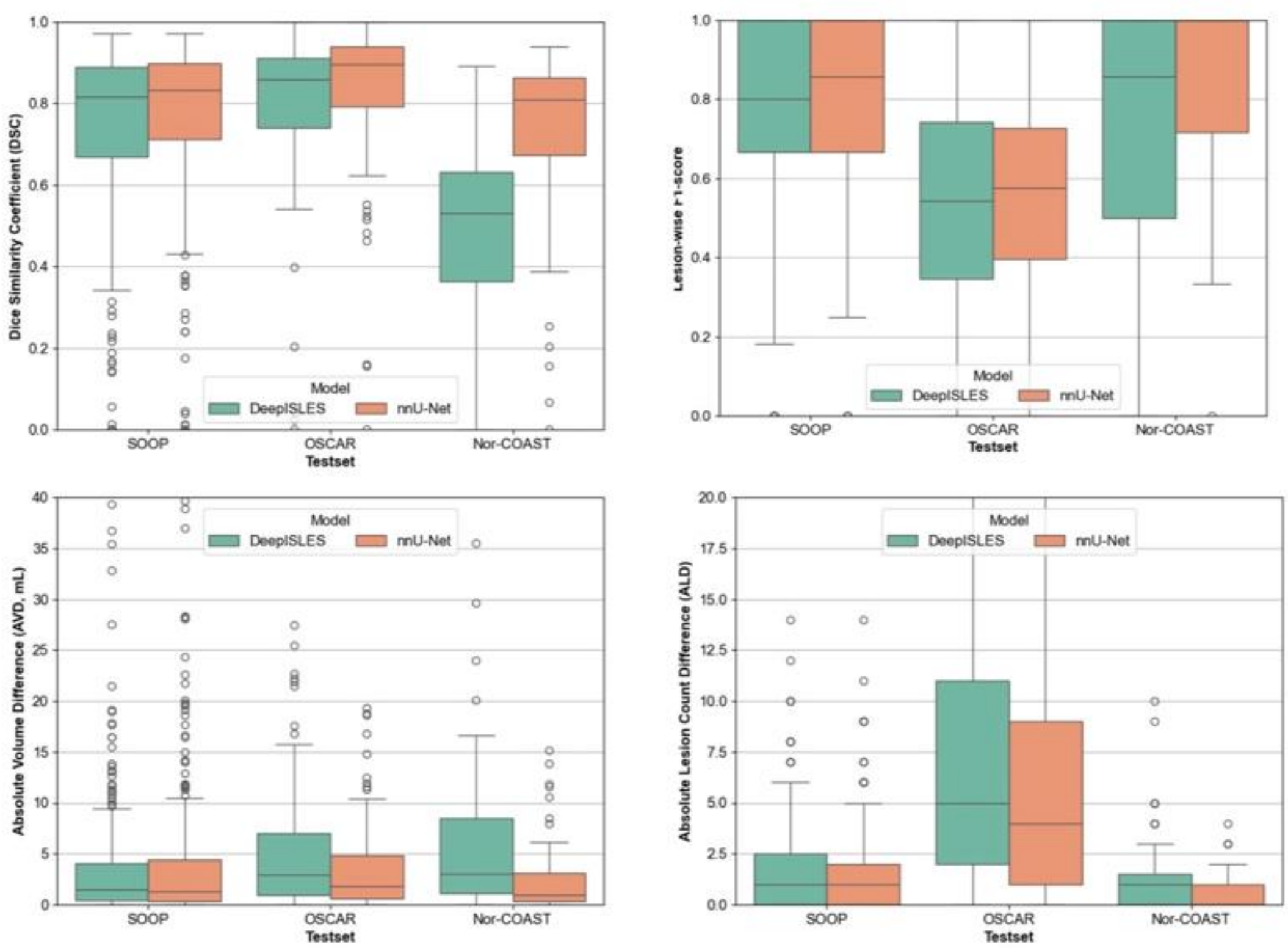

**Figure 2.** Boxplots comparing segmentation performance for the base model versus DeepISLES for the three test sets. SOOP = Stroke Outcome Optimization Project. OSCAR= Oslo Acute Reperfusion Stroke Study and Nor-COAST = Norwegian Cognitive Impairment after Stroke Study.

**Table 4.** Performance comparison between the base nnU-Net model and DeepISLES across the three test datasets.

| | SOOP (n=263) | | | | OSCAR (n=110) | | | | Nor-COAST (n=51) | | | |
|---|---|---|---|---|---|---|---|---|---|---|---|---|
| **Model** | **DSC ↑** | **F1 ↑** | **AVD ↓** | **ALD↓** | **DSC ↑** | **F1 ↑** | **AVD ↓** | **ALD↓** | **DSC ↑** | **F1 ↑** | **AVD ↓** | **ALD↓** |
| Base nnU-Net model | 0.83** (0.19) | 0.85** (0.33) | 1.29* (4.14) | 1.00** (2.00) | 0.90** (0.15) | 0.58 (0.34) | 1.85** (4.25) | 4.00** (8.00) | 0.81** (0.20) | 1.00* (0.31) | 0.96** (2.79) | 0.00* (1.00) |
| DeepISLES | 0.82 (0.23) | 0.80 (0.33) | 1.49 (3.72) | 1.00 (3.00) | 0.86 (0.18) | 0.54 (0.40) | 2.96 (6.30) | 5.00 (9.25) | 0.53 (0.29) | 0.86 (0.50) | 2.99 (8.07) | 1 (2.00) |

DSC = Dice Similarity Coefficient, F1 = Lesion-wise F1-score, AVD=Absolute lesion volume difference (mL), ALD= Absolute lesion count difference. Values are median (IQR). ↑↓ = higher/lower is better. Statistical comparison between models (Related-Samples Wilcoxon Signed Rank test): ** $p < 0.001$, *$p < 0.05$.

## Comparison of base nnU-Net versus DeepISLES models

In comparing the locally trained nnU-Net with the DeepISLES model, we opted for the base model because it gave a faster solution at inference and had comparable performance relative to the larger ResEnc. Table 4 and Figure 2 summarize the performance metrics for the base model relative to DeepISLES that was deployed without retraining. Across the three test sets, the median (IQR) DSC was 0.84 (0.19) for nnU-Net and 0.81 (0.19) for DeepISLES, which was a statistically significant difference in favour of the base model ($p < 0.001$). Hence, it showed non-inferiority or improved performance over DeepISLES across all metrics, and the difference was larger for the Nor-COAST cases where the median DSC was 0.81 for nnU-Net while DSC was 0.53 for DeepISLES. There was a significant ($p < 0.001$) positive correlation between GT- and predicted lesion volumes for both base and DeepISLES models, with a Pearson correlation coefficient of 0.978 and 0.957, respectively. Figure 3 shows Bland-Altman difference plots between GT and the DeepISLES stroke volume estimates, as well as GT versus the base nnU-Net model predicted stroke volume. Both DeepISLES and the base models tended to slightly underestimate the stroke lesion volume relative to GT. Figure 4 reveals that DeepISLES tended to produce poorer DSC score for stroke lesions that were 5 mL or smaller.

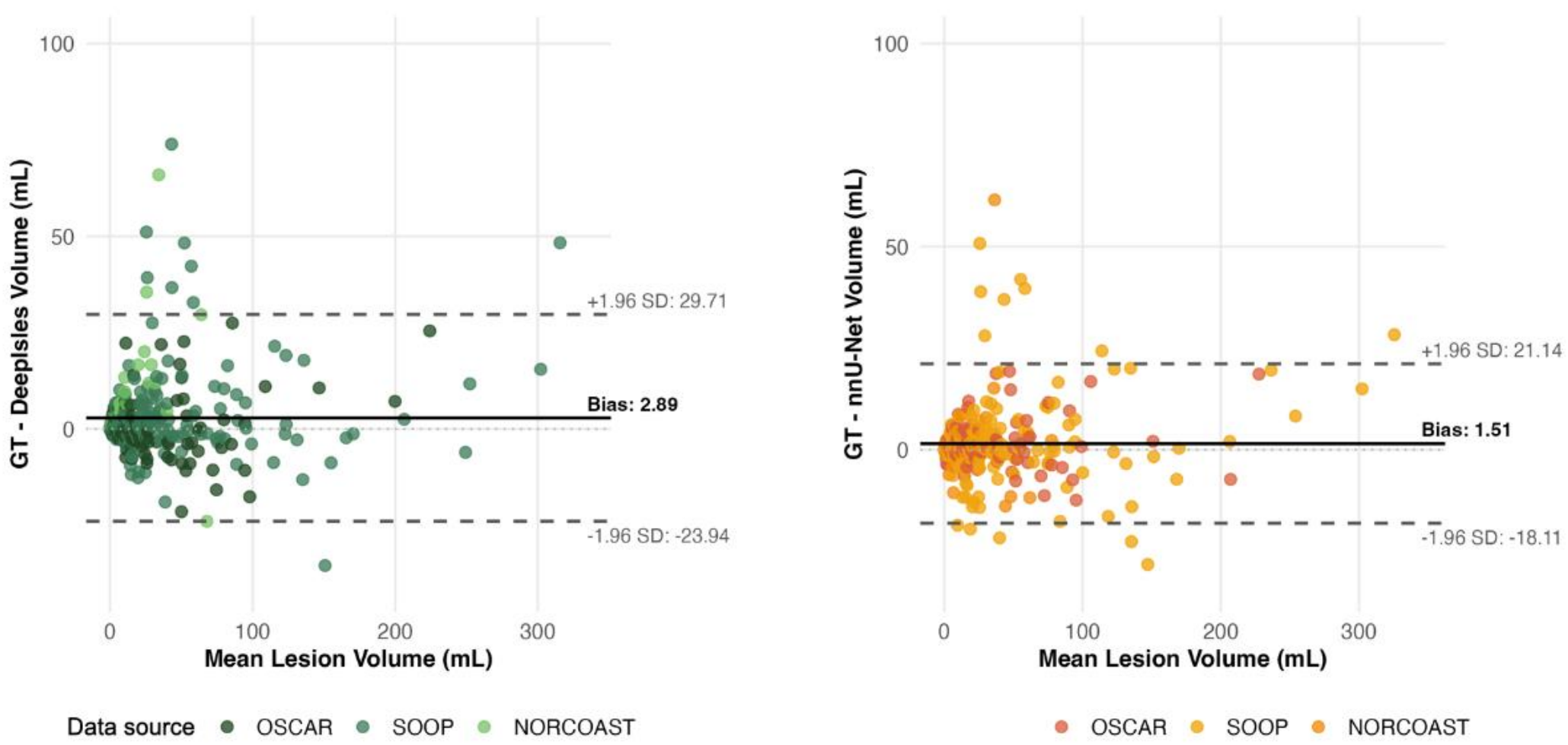

**Figure 3.** Bland-Altman plots show marginal differences between the models and ground truth (GT) stroke volume (mL) when considering the test samples (N=424). Left: ground truth minus DeepIsles segmentation prediction. Right: ground truth minus the base model prediction. The base model tended to show less stroke volume bias and had a lower standard deviation. A few cases from patients with large strokes and corresponding large discrepancies with the ground truth were excluded for visualization purposes. The colour corresponds to the different data sources.

The inference time per case was approximately 5 seconds and 235 seconds, respectively for nnU-Net and DeepISLES (including brain extraction).

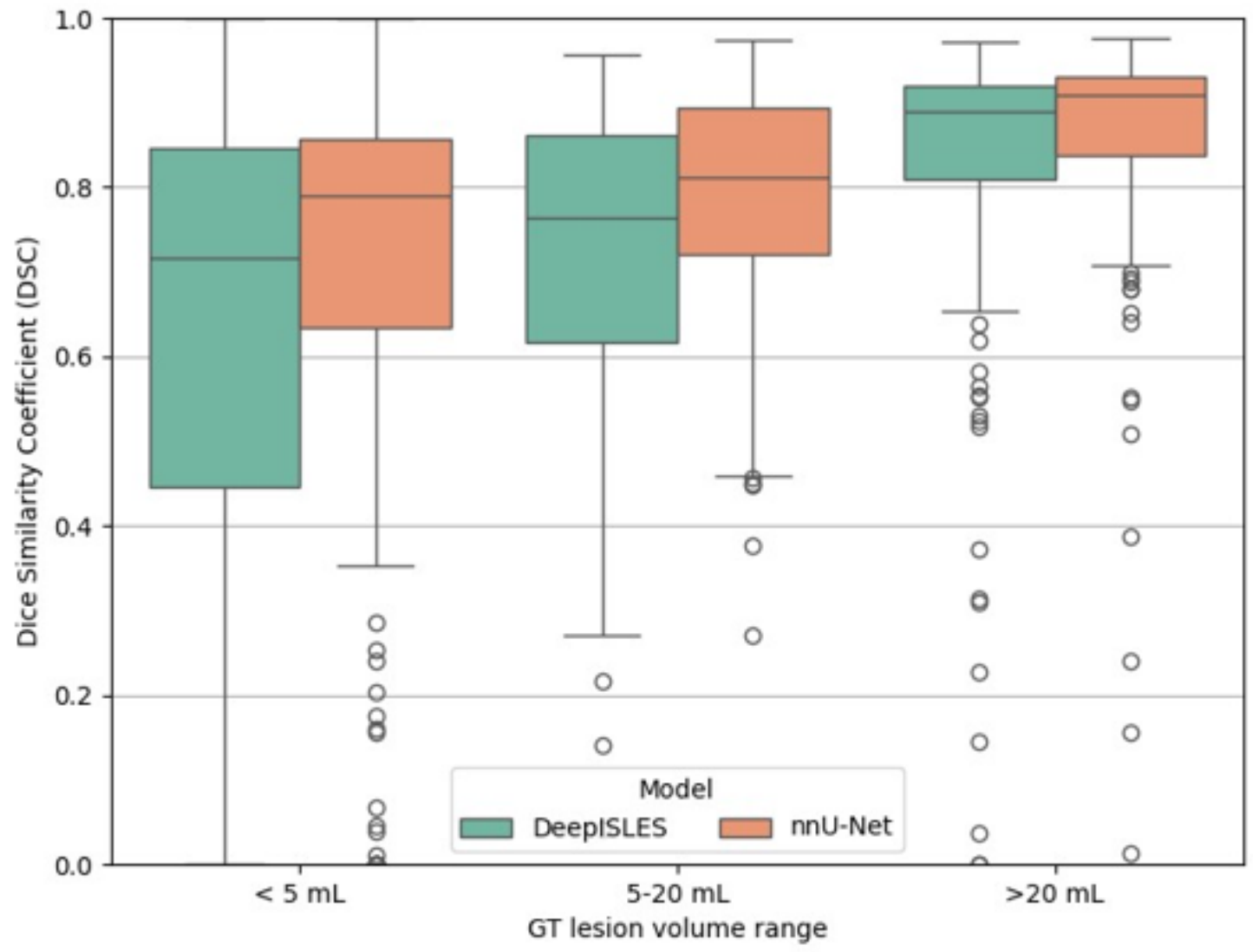

**Figure 4.** Comparison of Dice Similarity Coefficient (DSC) according to ground truth (GT) lesion volume range.

**Sample cases**

Figure 5 shows a patient with a large AIS from the OSCAR test set where both base and DeepISLES models achieved a DSC of 0.92. Figure 6 shows cases where the DeepISLES model failed to detect the lesion (DSC=0) whereas the base model correctly identified the lesion with a DSC of 0.77. Figure 7 shows a case from the Nor-COAST dataset of a patient presented with a small stroke lesion and large lesions from choroid plexus xanthogranuloma. Both the DeepISLES model and the nnU-Net model including ADC were able to distinguish between the small stroke lesion and the choroid plexus cysts. The base model using only the DWI image also correctly identified the stroke lesion but incorrectly interpreted the cyst-like lesion as stroke.

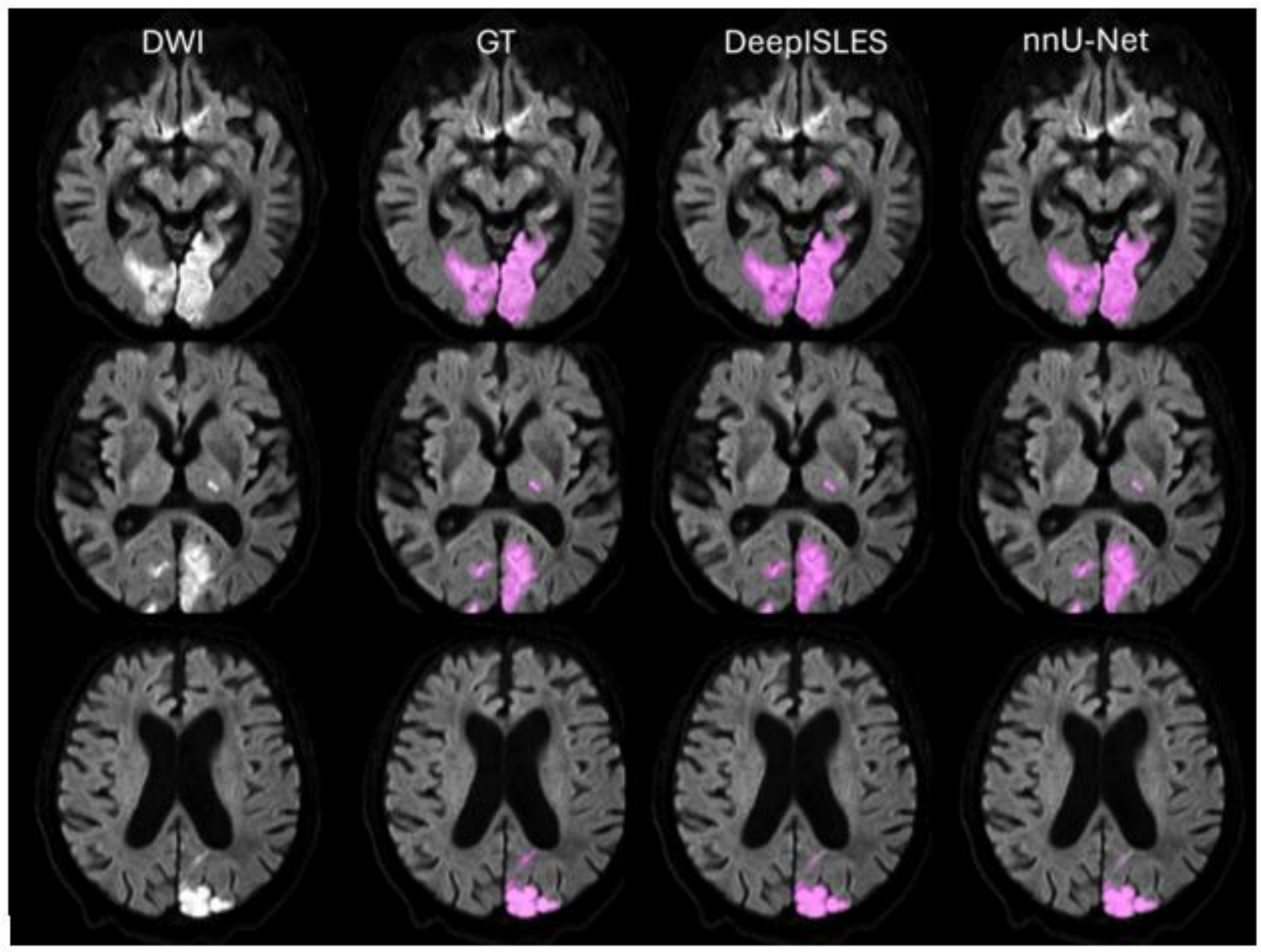

**Figure 5.** Sample case from the OSCAR test set with a large GT lesion volume of 52 mL where both models achieved a high DSC of 0.92. GT=ground truth lesion mask. DSC = Dice Similarity Coefficient.

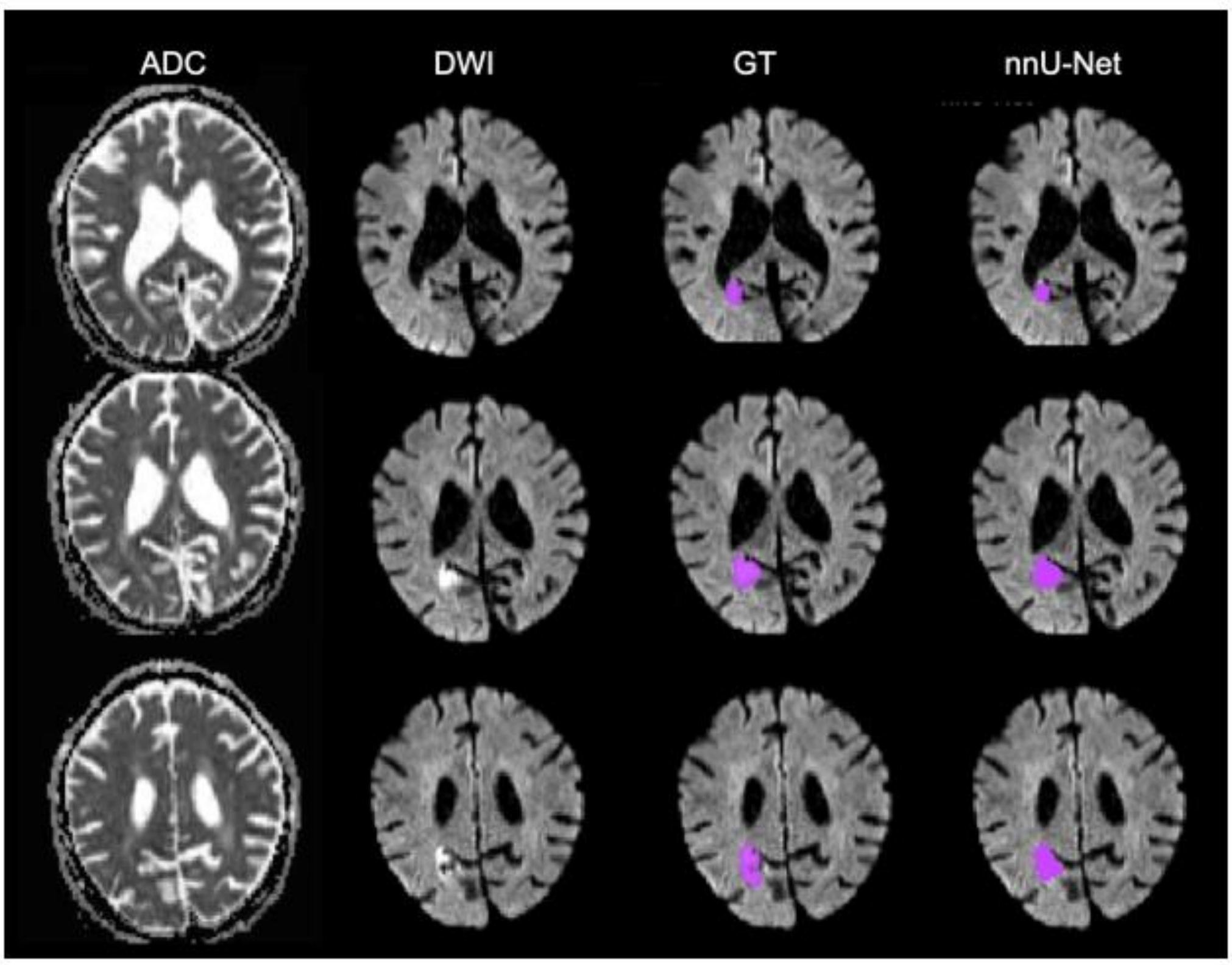

**Figure 6.** Sample case from the Nor-COAST test set where the DeepISLES model failed to detect the lesion (not shown in the figure) whereas the base model achieved a DSC of 0.77. GT=ground truth lesion mask. GT lesion volume = 3.6 mL. DSC = Dice Similarity Coefficient.

Figure 8 shows a case from the SOOP test set where GT annotation was deemed to be incorrect based on post hoc inspection, as the segmentation demarcated a hyperintense region in the choroid plexus and not in brain parenchyma. Neither base / ResEnc nnU-Net or DeepISLES models identified this as an AIS lesion.

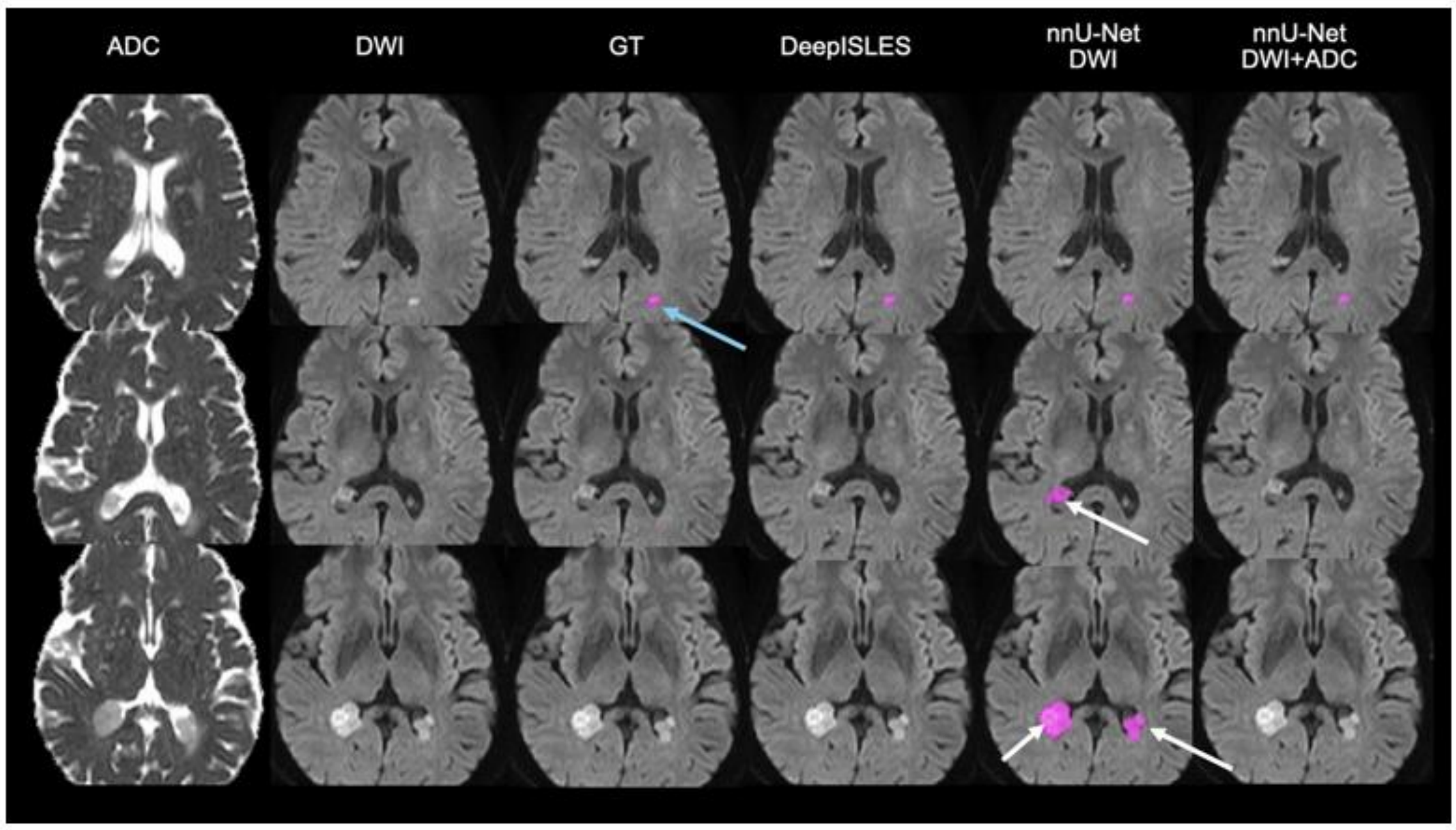

**Figure 7**. Sample case from the Nor-COAST test set of a patient presenting with a small stroke lesion (blue arrow) combined with multiple large lesions due to choroid plexus xanthogranuloma (white arrows). Whereas both the DeepISLES model and the nnU-Net model including ADC as input channel were able to distinguish the cystic lesions from stroke, the nnU-Net trained only on DWI failed to distinguish the two lesion types. GT = ground truth lesion mask.

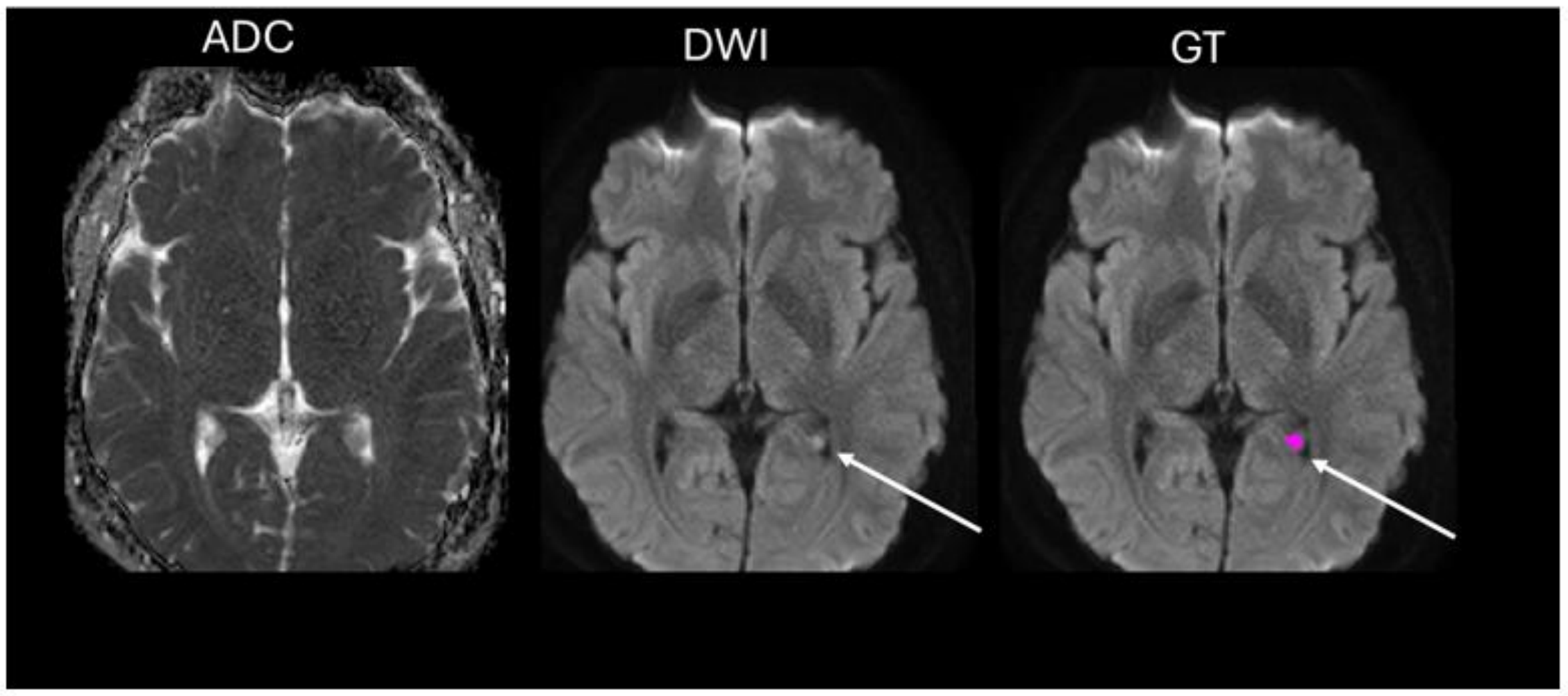

**Figure 8.** Sample case from the SOOP test set where both models (DeepISLES and baseline nnU-Net) correctly ignored the small cyst-like lesion in the choroid plexus (white arrow), which was an incorrect GT annotation. GT=ground truth lesion mask.

## Discussion

The current study investigated automated DWI segmentation with findings summarized by the following. First, the larger ResEnc nnU-Net model with residual encoder connections had slightly better performance than the base model in terms of Dice score. There was however no difference in performance between these models when considering F1-score, absolute lesion volume, or absolute lesion count metrics. The marginal improvement in the ResEnc model comes with a significant compute cost owing to the larger model and translates to longer inference times and memory use. Second, the base nnU-Net model was not inferior to the DeepISLES ensemble model; in contrast, it significantly outperformed the established DeepISLES model under certain experimental conditions. Third, visual inspection of segmentation results revealed segmentation errors for patients with unexpected incidental or co-pathologies alongside the AIS. Collectively, the findings capitalize on well-curated, diverse, and annotated DWI to reinforce the value of training deep learning segmentation models by using local data in addition to or instead of relying on open-source data.

The current findings revealed that skull-stripping the DWI, a method that is standard for structural T1-weighted MRI, did not meaningfully improve segmentation results. Although removing non-brain voxels is sensible, it introduces a longstanding challenge of DWI signal dropout that is corrected by additional scans that are not clinically routine (23). Previous studies investigated this by considering single and multiple channel model inputs. Kamel et al, however, did not observe a segmentation performance gain when ADC and/or FLAIR were added as model inputs, but a brain extraction step was not considered (24). Skull-stripping does not seem to be warranted for DWI stroke segmentation and potentially introduces clinical implementation issues due to inaccuracies in the brain extraction (e.g. under or overestimating the brain tissues), the need for additional image analysis steps, and the reliance on additional sequences.

In terms of the raw DWI and the corresponding derived images, a case could be made to include the ADC map to mitigate segmentation inaccuracies for stroke mimic cases and other pathologies. We found that the models trained with ADC maps as a second input channel tended to protect against false positive errors associated with T2 shine-through (e.g. multiple choroid plexus xanthogranuloma lesions). The T2 shine-through is caused by high DWI signal due to elevated T2-relaxation times, and not restricted diffusion (22). Adding ADC as model input may then aid in differentiating strokes from causing high DWI signal but not associated with restricted diffusion. It is noted that we observed a single case of T2 shine-through from among 436 cases, which was correctly segmenting by the DWI + ADC two channel model.

Compared to the DeepISLES ensemble model in the same held-out test sets, the nnU-Net compared favourably across all performance test metrics. Median DSC was 0.84 and 0.81, respectively for nnU-Net and DeepISLES, the value for DeepISLES being in close agreement with a previous report in acute and early sub-acute ischemic stroke cohort (7). Notably, DeepISLES was significantly worse for the Nor-COAST sample with a DSC of 0.53 compared to 0.81 for nnU-Net. Inspecting the predictions on a case-by-case basis, we observed that DeepISLES failed to detect any lesions in seven cases (out of 51) from Nor-COAST cohort and nnU-Net only failed in one of these cases. Although the DeepISLES model previously showed similar segmentation performance on test data from an external imaging center, unseen during model training, the result in our study may indicate that the images in the Nor-COAST dataset deviated significantly in terms of image quality or contrast from the DeepISLES model training data. Figure 5 shows a sample case from the Nor-COAST where DeepISLES failed to detect the lesion, but this also occurred in data from the other test cohorts, and comparing segmentation performance according to lesion size, we found that DeepISLES generally performed worse in cases with smaller lesion volumes (Figure 4). This highlights the challenge of maintaining robust segmentation performance in out-of-sample data.

The current study is not without limitations that should be discussed. We did not include an out-of-sample test set from a cohort that was not seen by any of the models during training. However, the main purpose of this work was to validate a segmentation tool meant for clinical use within our hospital network and regionally, thus sufficient performance in relevant data acquired from our institutions was therefore considered more important than showing robust performance across the widest possible range of image qualities and protocols. Furthermore, since we aimed to use native DWI data without preprocessing in terms of brain extraction most open-source datasets are already brain extracted and therefore not useful for this specific test. For this reason, we included data from the Stroke Outcome Optimization Project, as a large-scale open-source dataset that had unprocessed DWI and ADC data. The combination of this large dataset with both a national dataset from five institutions (Nor-COAST) and a hospital-specific dataset from our institution (OSCAR) still ensured training data from a wide range of scanners and imaging protocols.

In conclusion, we found that use of the nnU-Net with a single DWI input and no preprocessing enabled fast and accurate AIS lesion segmentation; this approach may facilitate efficient clinical deployment in acute stroke imaging workflows.

**Acknowledgments**

This work was performed on a high-performance computing cluster named Fox (Educloud Research) that is provided by the University of Oslo. We thank the steering committee for granting access to the Nor-COAST MRI data. Nor-COAST is a collaboration between the Norwegian Health Authorities, five hospitals and Universities in Norway represented through a steering committee with project leadership at Department of Neuromedicine and Movement Science, NTNU, Trondheim. Participating partners are Central Norway Regional Health Authority: St. Olav University Hospital, Ålesund Hospital, Møre and Romsdal Health Trust; South-Eastern Norway Regional Health Authority: Vestre Viken Hospital Trust, Bærum Hospital, Department of Radiology and Nuclear Medicine Oslo University Hospital, Department of Geriatrics, Oslo University Hospital; and Western Norway Regional Health Authority: Haukeland University Hospital.